\documentclass[journal]{IEEEtran}

\ifCLASSINFOpdf
\else
   \usepackage[dvips]{graphicx}
\fi
\usepackage[american]{babel}
\usepackage{cite}
\usepackage{url}
\usepackage{multirow}
\usepackage{caption}
\usepackage{subcaption}
\usepackage{amsfonts}
\usepackage{booktabs}
\usepackage{amsmath}
\usepackage{float}
\usepackage[switch]{lineno}

\usepackage{lineno,hyperref}
\usepackage{xcolor}
\usepackage{graphicx}
\begin{document}

\title{Where is the Model Looking At? \\  --
Concentrate and Explain the Network Attention.}

\author{Wenjia Xu, Jiuniu Wang, Yang Wang, Guangluan Xu, Wei Dai, and Yirong Wu

\thanks{}
\thanks{Wenjia Xu, Jiuniu Wang, Yang Wang, Guangluan Xu, Wei Dai, and Yirong Wu are with Key Laboratory of Network Information System Technology (NIST), Institute of Electronics, Chinese Academy of Sciences,  (e-mail: xuwenjia16@mails.ucas.ac.cn).}}

\markboth{IEEE JOURNAL OF SELECTED TOPICS IN SIGNAL PROCESSING, VOL. 14, NO. 3, MARCH 2020}
{XU \MakeLowercase{\textit{et al.}}: WHERE IS THE MODEL LOOKING AT? – CONCENTRATE AND EXPLAIN THE NETWORK ATTENTION}
\maketitle

\begin{abstract}
Image classification models have achieved satisfactory performance on many datasets, sometimes even better than human. However, The model attention is unclear since the lack of interpretability. This paper investigates the fidelity and interpretability of model attention. We propose an Explainable Attribute-based Multi-task (EAT) framework to concentrate the model attention on the discriminative image area and make the attention interpretable. We introduce attributes prediction to the multi-task learning network, helping the network to concentrate attention on the foreground objects. We generate attribute-based textual explanations for the network and ground the attributes on the image to show visual explanations. The multi-model explanation can not only improve user trust but also help to find the weakness of network and dataset. Our framework can be generalized to any basic model. We perform experiments on three datasets and five basic models. Results indicate that the EAT framework can give multi-modal explanations that interpret the network decision. The performance of several recognition approaches is improved by guiding network attention.
\end{abstract}


\IEEEpeerreviewmaketitle

\section{Introduction}

The deep neural network has gained significant achievements in many computer vision tasks. Despite its superior performance, the complex network is lack of interpretability. The intelligent cannot explain the causes of their behavior, and provide no explanation for either the superior performance or unsatisfactory result. As a consequence, users would not know if the network is trustworthy even though they achieve high precision.

\begin{figure}
    \centering
    \includegraphics[width=1\linewidth]{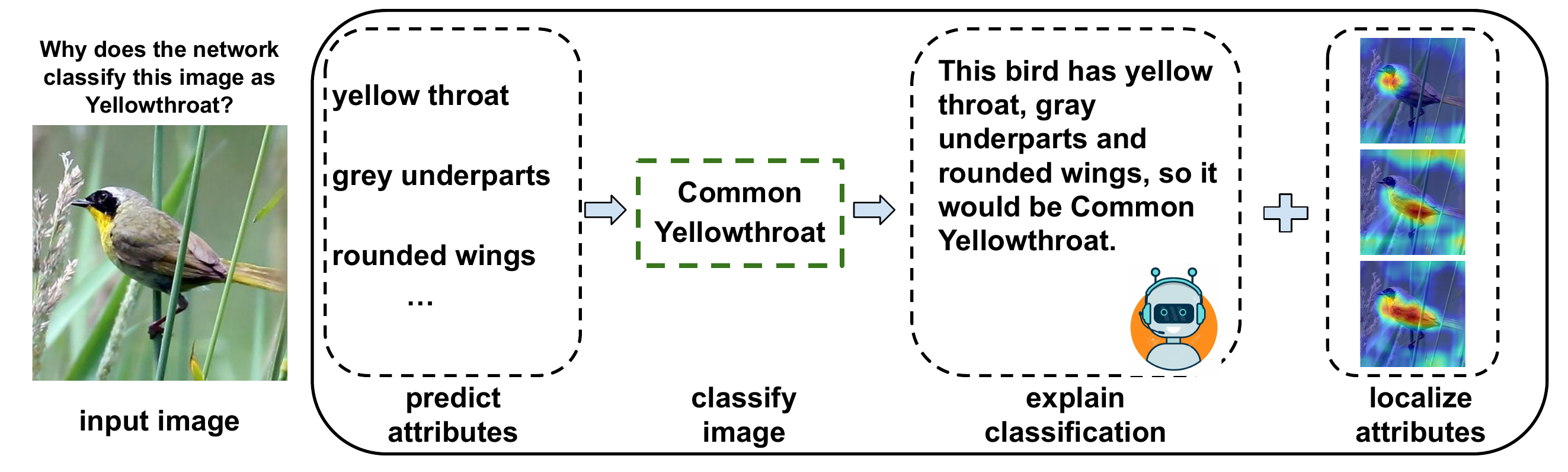}
    \caption{
        With an input image, we first predict the attributes of the foreground object, then classify image category according to the attributes. By calculating the attribute contribution to the classification result, we can conclude which attributes influence the network decision. We explain the classification network with those important attributes and ground the attributes on the input image.
    }
    \label{fig:teaser_figure}    
\end{figure}

For many real-world deep learning tasks, only getting a high prediction accuracy is not enough. Many researchers are trying to make the network more stable and interpretable. This paper addresses these two problems by concentrating the network attention and giving multi-model explanations. More specifically, given an input sample, we build a framework that can focus on the discriminative foreground object. Moreover, we provide a textual-visual combined explanation to interpret the model. An output example is shown in Figure~\ref{fig:teaser_figure}.

\begin{figure}
        \centering
        \includegraphics[width=1\linewidth]{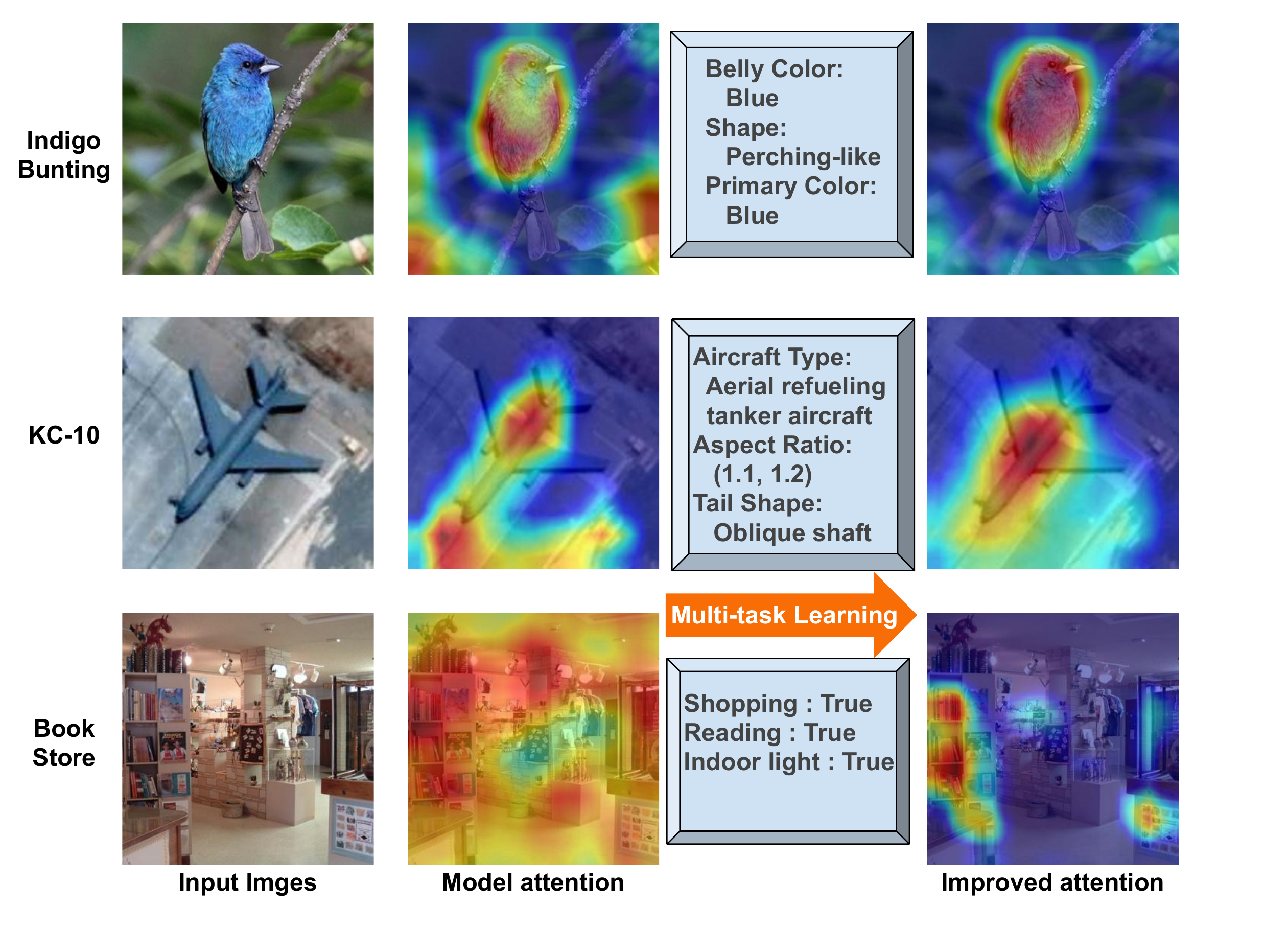}
        \caption{
            The proposed multi-task framework helps the network attention to focus on the necessary objects instead of the background. The second column shows the original attention map of a classification network based on ResNet50~\cite{he2016deep}, which distributes attention on the background such as leaves, airport runway and groceries in the store. The third column lists the most discriminative attributes that help the network to make a prediction. The important attributes are predicted by EAT framework. The attention maps on the fourth column are from EAT framework. With the help of attribute prediction, the network attention is guided toward the core object of the classification tasks.
        }
        \label{fig:back_ground}    
\end{figure}

To deal with these problems, the first question to be raised is ``Is the network paying attention to the right part when making a decision?'', , or equivalently, ``How to concentrate the network attention on foreground objects?''. 
One interesting phenomenon is observed in \cite{20_grad_cam} and \cite{li2018tell} that the data bias might influence network attention. For example, in image recognition tasks, when the foreground objects always come along with a similar background, the training network might classify images partly according to the background. In Figure~\ref{fig:back_ground}, the second column shows some biased network attention. When classifying an image labelled bird ``Indigo Bunting'', the classification network pays attention to the leaves and branches. However, the performance of this over-fitting network will suffer in testing data with various backgrounds, i.e. a bird flying in the sky or floating on the water.

When recognizing an image, a human would map the foreground object into an attribute space, then classify the image according to the most discriminative attributes. Similarly, we take advantage of the attributes, to lead the network attention on the foreground image area. In this paper, we propose an explainable attribute-based multi-task learning framework (EAT) to map the image pixels both into attributes space and category space. The multi-task framework shares the parameters for predicting the attributes label and the bird category. When performing attribute prediction, the network is taught to focus on the discriminative parts on the foreground. In this case, the model would be convinced about which feature is relevant to the whole task and focus on them.

As shown in Figure~\ref{fig:back_ground}, when classifying the airplane, the attention map in EAT framework are focused on the object, instead of the runway. When classifying book store, the network can pay more attention to the vital factor, the books, instead of the background environment.

Leading the network to be sensitive to the discriminative parts would make it more stable. After that, the second question to be raised is ``How to explain the network attention to users?'' and ``How to make the network more interpretable and trust-worthy?''. In this paper, we generate intuitive explanations to show the network decision process. Since the category prediction is inferred by the attribute distribution of input image, we can explain the network by stating which attribute contribute more to the output.

In the EAT framework, we propose an attention embedding reasoning (EAR) module to calculate the importance of every attribute on the output. Then, attributes with higher contribution would form the textual explanation. As shown in Figure~\ref{fig:teaser_figure}, our framework generates attribute-based explanation to answer the question ``Why does the network classify the image as Yellowthroat?''.
Besides, attributes provide discriminative localization information which can be located on the image, e.g., the yellow throat and grey belly of bird ``Yellowthroat''. Thus we can visually ground the attributes with attention map, providing a textual-visual combined explanation.

To evaluate the effectiveness of EAT framework, we perform experiments on fine-grained image recognition task which aims to distinguish sub-categories. Since sub-categories are similar in general appearance, the introduce of attributes will guide the machine to find subtle and local differences. The multi-modal explanations can help users to better understand the network predictions, both the success and failure cases. Note that our framework can also be performed on other tasks. We also do experiments on large-scale scene recognition task~\cite{25_SUNdataset}.

The main contributions are summarized as follows:
    \begin{itemize}
        \item{We propose an attribute-based multi-task framework (EAT) that integrates attribute prediction with image classification. The framework provides discriminative information to classify images. Meanwhile, the introduction of attribute prediction helps the network to pay attention to relevant features on the foreground object rather than the background.
        }
        
        \item{To make the decision process more interpretable, we propose an embedding attention reasoning (EAR) module to reveal the important attributes that are responsible for the model prediction. We calculate the attribute contribution to the classification procedure, thus generate the attribute-based language explanation. The essential attributes can also be visually grounded on the image via model attention, providing multi-modal explanations for the network attention.}
        
        \item{Our framework can be performed on basic classification networks such as Alexnet~\cite{Alexnet}, ResNet~\cite{he2016deep}, and well-designed fine-grained recognition networks such as DFL~\cite{27_DFL}. Experiment results indicate that the EAT framework can also help models to focus on the foreground objects that provide subtle differences for classification. Our framework can also generate explanations that make the network attention more interpretable.  }
        
    \end{itemize}
    
\section{Related Work}
\subsection{Explaining Neural Network}

Explaining deep neural networks has been extensively studied in recent years. Interpretation for the network would help users to understand the model and the data. The interaction between network and users also improve user trust and help them to debug the model. Previous interpretation are applied on image classification~\cite{hendricks2016generating, 22_zhang2018top, 19_cam}, visual question answering~\cite{20_grad_cam}, automatic driving~\cite{kim2018driving} and so on. 

Interpretation methods can be differentiated by various criteria. For instance, whether the model is explained directly (intrinsic), or after training (post hoc); whether the interpretation can explain a specific prediction (local) or explain the behavior of the whole model (global). In this work, we mainly focus on post hoc and local explanations that interpret a single prediction of a trained network. According to the form of explanation, they can be classified to visual explanation that provides attention maps~\cite{22_zhang2018top, 16_deconv, n14, n13_zelier}, and textual explanation that generates sentences~\cite{10_su2017reasoning,hendricks2016generating,hendricks2018grounding,kim2018driving}.  Visual explanation methods generate attention maps to highlight the important image regions that influence the network decision. CAM~\cite{19_cam} and Grad-CAM~\cite{20_grad_cam} backpropagate the prediction score to a particular feature map, then generate attention map according to the important feature maps. LIME~\cite{21_lime} and RISE~\cite{18_rise} regard the model as a black box, generate masks to cover the input image and determine important image area by the prediction score for masked images. Users can figure out and solve some restrictions of the training network with the help of visual explanations.

Except for explaining the network decision with attention maps, textual explanations use natural language to interpret the network output. Hendricks et al.~\cite{hendricks2016generating} feed images and the classification results to LSTM~\cite{38_LSTM}, generating textual descriptions that point to the evidence. Huk Park et al.~\cite{huk2018multimodal} provide language description to explain the visual question answering network. Kim et al.~\cite{kim2018textual} interpret the behavior of a self-driving system. However, training LSTM requires human-annotated descriptions for every image, which is time-consuming. Due to the data bias of the training descriptions, it is prone to generate sentences that are not related to the image and classification results. Hendricks et al.~\cite{hendricks2018grounding} propose an evaluation metric to measure how well is an explanation sentence related to the image. However, the causal relationship between prediction results and the explanations is still weak. In this work, we generate more accurate textual explanation by calculating how much does every attribute influence the classification result. Besides, the attribute-based explanation can be grounded on the image via attention maps, providing a multi-modal explanation for every single prediction.

\subsection{Attribute}

Visual attributes are additional annotations that describe the semantic properties of objects and scenes (e.g., shape, color and location). When classifying birds, for example, attributes can describe a bird that has ``white belly'', ``long leg'' and ``middle size'', etc. Since attributes are both machine-detectable and semantically meaningful, learning with attributes has been explored for various applications. Most notably, zero-shot learning establishes the relationship between attributes and images, and build a model that classify images without seeing any training examples ~\cite{1_lampert2009learning, 2_xian2018zero,3_xian2018feature}. Attributes also enable innovative applications like face recognition~\cite{7_kumar2009attribute, 8_hu2017attribute}, and image-to-text generation such as image caption~\cite{6_zhang2014panda,18_rise}. 

In our work, the attributes are used to help the network to focus concentration and generate explanations. Note that we need class-attributes to train the multi-task learning network and generate textual explanations. Different from bounding box or language descriptions that require considerable effort for annotation, the class properties for nearly all the objects are already defined. Thus, it would be quick and cheap to label class-attributes~\cite{32_awa}. 

\subsection{Improving Model Concentration}
As shown in Figure~\ref{fig:back_ground}, when classifying an aircraft labelled ``KC-10'', the network lays some attention on the runway. This is caused by the training data bias, where the foreground object is always correlated with a similar background. The data bias between two datasets can be measured by training a model on one dataset and testing it on another dataset~\cite{torralba2011unbiased}. There are two methods to solve this data bias problem. The first solution is to enlarge the amount of training data. 
However, collecting data is time-consuming, and it is hard to remove all the biases. The second solution is to improve the model generation ability. Recently, many methods have been proposed to tackle this problem. 
With the help of segmentation labels, the guided attention inference network~\cite{li2018tell} trains the network with two loss
functions, to guide the network focus on the objects. Some unsupervised representation learning methods use image rotation prediction \cite{gidaris2018unsupervised} or spatial context prediction \cite{doersch2015unsupervised} to help training network focus attention on foreground objects.

In our work, different from the existing methods, we deal with this problem naturally. When classifying the input image, the neural network treats every pixel equally, and try to map the pixels to a label space. If the network is trained on a biased dataset where the target object often comes with a similar background, the network might overfit on the dataset by paying attention to the background. However, if the model is jointly trained with other tasks that focus on the same forehead object, these tasks would provide additional evidence on which feature is essential. Thus the network attention would be focused on the features that matter.

\subsection{Multi-task Learning Network}
In machine learning, researchers usually propose a single model to deal with their main task. However, there is some information from other tasks that can help to improve the metric. Multi-task learning (MTL) is a solution that joint train several tasks~\cite{ruder2017overview}, and share the domain-specific information contained in them. MTL has achieved success across many deep learning applications~\cite{deng2013new, ramsundar2015massively, long2015learning}. If the training data is noisy or limited, it would be hard for the model to distinguish relevant features from the data. While MTL would jointly train the model on other tasks, and improve the generalization of the main task by helping the model to pay attention to the features that matter.

There are two types of MTL methods, soft parameter sharing and hard parameter sharing.  In soft parameter sharing, different tasks have their own network. To make the networks similar, their parameters are regularized by a distance function. In hard parameter sharing methods, different tasks share the same hidden layer, while having a specific output layer. 

Our EAT framework is designed following hard parameter sharing. We share the same feature extraction network for attribute prediction and image category prediction. Jonathan shows in~\cite{baxter1997bayesian} that multi-task learning can prevent the model from overfitting on the main task. It is proved in our experiment that when the network captures the common features for attributes classification and category prediction, the chance of overfitting is low. Especially for our fine-grained recognition dataset CUB~\cite{26_wah2011caltech} and Aircraft-17~\cite{aircraft17}, where the birds often come with a similar background, such as forest or water, a single classification network are prone to make a decision according to the background. This network will fail on an image with an unusual background. Since attribute prediction captures the characters of objects, joint training the hidden parameters of attribute prediction and category classification will prevent the network from overfitting.

\begin{figure*}[!t]
		\centering
		\includegraphics[width=1\textwidth]{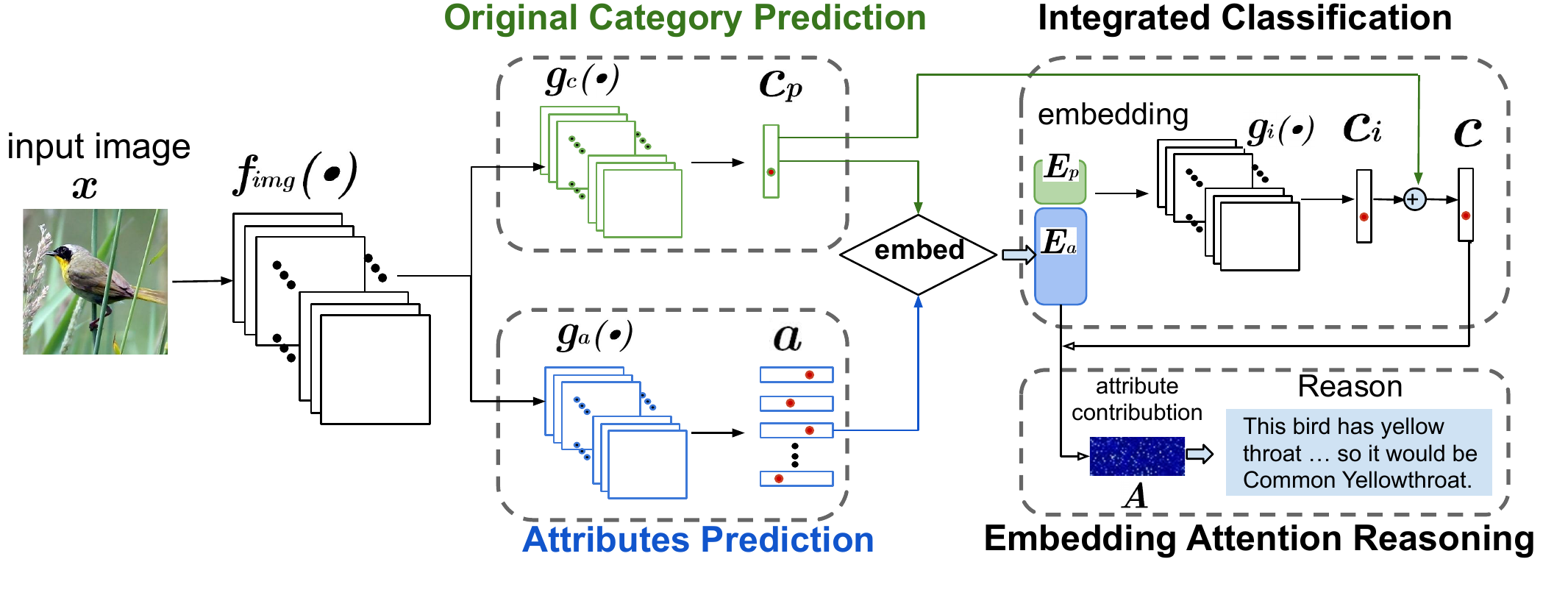}
		\caption{
			Overview of our explainable attribute-based multi-task framework (EAT). We split the base model into two parts, the feature extraction network $f_{img}(\cdot)$ and the original category prediction network $g_{c}(\cdot)$. In multi-task learning framework EAT, we construct three additional modules, i.e., attributes prediction, integrated classification, and embedding attention reasoning.
            In attribute prediction, $g_{a}(\cdot)$ predictes the label $a$ for every attribute. In integrated classification, $g_{i}(\cdot)$ is used to classify the embedding of $c_{p}$ and $a$, to obtain integrated category predicted result $c_{i}$. Then $c_{p}$ and $c_{i}$ is combined to obtain the final classification result $c$. In embedding attention reasoning, we obtain the attributes contribution map $A$ by back-propagating $c$ on attributes embedding $E_{a}$. The most important attributes are selected to form the classification reason.
		} 
		\label{fig:structure}    
\end{figure*}

\section{Explainable Attribute-based Multi-task Framework}
\label{sec:model}
In this section, we introduce how to integrate attribute prediction with image recognition, and generate explanations to interpret the network. This framework can be performed on classical image category prediction networks such as  AlexNet~\cite{Alexnet}, ResNet~\cite{he2016deep}, PnasNet5~\cite{liu2018progressive} and well-designed models like DFL~\cite{27_DFL}. As illustrated in Figure~\ref{fig:structure}, except for the original category prediction module, our mechanism consists of three additional modules, i.e., attributes prediction module, integrated classification module and embedding attention reasoning (EAR) module. 

    
\subsection{Attributes Prediction}
Given an image $x$, an original category prediction module aims to predict the correct category $c$. A general end-to-end model can be divided into two parts, feature extraction network ${f_{img}(\cdot )}$ which extracts features from the input image, and category prediction ${g_{c}(\cdot)}$ which maps the feature to category label:

\begin{equation}
\begin{aligned}
& {{v}_{img}}={{f}_{img}}(x) \\ 
& {c_p}={{g}_{c}}({{v}_{img}}) \\ 
\end{aligned}
\end{equation}
where ${v_{img}}$ denotes the image feature extracted by ${f_{img}(\cdot)}$, and $c_p \in {{\mathbb{R}}^{1 \times {{N}_{c}}}}$ is the output of an original category prediction network. ${N}_{c}$ denotes the number of image categories. Note that ${f_{img}}$ and ${{g}_{c}}$ can be replaced by any end-to-end image recognition networks.

In the attributes prediction module, the class attributes are introduced to the general end-to-end model, helping the network to focus on discriminative location information. We map image features ${v_{img}}$ to the class attributes labels $a_i, i \in {[1, N_a]}$. For each attribute, we apply a classifier ${{g}_{a}^i}$:
\begin{equation}
{a_i}={{g}_{a}^i}({{v}_{img}})
\end{equation}
where $a_i$ denotes the predicted result of the $i$-th  class attributes. ${g_{a}}$ denotes the attribute classifiers, in which each attribute has a corresponding classifier ${g_{a}^i}$. For simplicity, we utilize the same architecture for ${{g}_{c}}$ and ${{g}_{a}^{i}}$.

In our multi-task learning framework, the original category prediction network and attribute prediction network share the parameters of ${f_{img}(\cdot)}$. In this case, training two networks together would help ${f_{img}(\cdot)}$ to find what feature is relevant to both tasks. For instance, the attribute prediction network would only focus on the bird, because the background cannot provide any information for a specific attribute. Thus the original category prediction network will be guided to pay attention to foreground objects.

\subsection{Integrated Classification}
In order to take advantage of the vital location information extracted by the attribute prediction, we integrate the predicted results of class attributes $a$ and original category $c_p$, and feed them to an integrated classification network. 

In detail, we first align the dimension of ${a}$ and ${c_p}$ by embedding them into a $D_e$-dimensional vector:
    
\begin{equation}
\begin{aligned}
& {{E}_{a}}=em{{b}_{a}}({{a}}) \\ 
& {{E}_{p}}=em{{b}_{p}}({{c}_{p}}) \\ 
\end{aligned}
\end{equation}
where ${{E}_{a}}\in {{\mathbb{R}}^{{{N}_{a}}\times {{D_e}}}}$ is the class attributes embedding, and ${{E}_{p}}\in {{\mathbb{R}}^{1\times {{D_e}}}}$ is the preliminary category embedding.
    
Then, we concatenate ${E}_{a}$ and ${E}_{p}$ to get integrated embedding ${E}$ as
    
\begin{equation}
E=[{{E}_{a}};{{E}_{p}}] \,.
\end{equation}
We utilize a 3 layers CNN classifier ${{g}_{i}}(\cdot )$ to map the embedding $E$ into image category. 
\begin{equation}
\begin{aligned}
& {{c}_{i}}={{g}_{i}}(E) \,.
\end{aligned}
\end{equation}
At last, the final image category predicted result $c$ is the weighted sum of $c_p$ and $c_i$. 
\begin{equation}
\begin{aligned}
& c=\lambda \cdot{{c}_{p}}+\eta \cdot {{c}_{i}} \\ 
\end{aligned}
\end{equation}
where $\lambda$ and $\eta $ are the hyper parameters that control how much should the attribute prediction influence the image category prediction.

In the training process, cross entropy $CE(\cdot)$ is applied as the objective function, aiming to minimize the loss for integrated category prediction and attribute prediction:
\begin{equation}
\begin{aligned}
&{{l}_{c}}=CE({y},{{c}}) \\
& {{l}_{a}}=\frac{1}{{{N}_{a}}}\sum\limits_{j=1}^{{{N}_{a}}}{CE({\cal A}_{i},a_i)} \\ 
\end{aligned}
\end{equation}
where ${\cal A}$ denotes the ground truth of class attributes, and ${y}$ denotes the ground truth of the image category. 
Here 
\begin{equation}
CE(gt,p)=-\frac{1}{D}\sum\limits_{i=1}^{D}{{{gt}^{(i)}}\log }({{p}^{(i)}})
\end{equation}
where $gt$ is the ground truth label, $p$ is the predicted probability, and $D$ is the dimension of $gt$ and $p$.

\subsection{Embedding Attention Reasoning}

For every class, there are hundreds of attributes describing all the properties of the objects. To figure out which attributes are essential when classifying the image, we calculate the attribute contribution to the network prediction in this module, then generate the attribute-based explanation for the network decision. 

As is shown in Figure~\ref{fig:structure}, in EAR module, we backpropagate the final prediction $c$ to the attribute embedding $E_a$, get the gradient of $c$ on the attributes embedding $E_a$:
\begin{equation}
W =\frac{\partial c}{\partial E_a}
\end{equation}
where $W \in {{\mathbb{R}}^{{N_a}\times {{D_e}}}}$ indicates the network attention over the attribute embedding. The blue star map shown in Figure~\ref{fig:structure} is the visualization results of $W$, where each line $W_i$ represents the attention value of a corresponding attribute.

The attention contribution score $s_i$ for the $i$-th attribute embedding is the accumulation of $W_i$,
\begin{equation}
{{s}_{i}}=\sum\limits_{j=1}^{D_e}{{{W}_{ij}}}.
\end{equation}

Those attributes with the highest contribution score are contributing more to the final prediction result. As shown in Figure~\ref{fig:teaser_figure}, ``Yellow Throat'', ``Grey Underparts'' and ``Rounded Wing'' are the most important attributes that help the model to make a decision. Then the top-3 attributes with the highest attention value are summarized into a textual explanation, to indicate where is the model attention when performing classification.


\section{Experimental Setup And Evaluation Metrics}

In this section, we present the datasets and the environmental setting. Then we introduce how to generate the network attention maps. Afterwards, we propose the metric to calculate how much is the network focusing on forehead objects.

\begin{table}
    \begin{center}
        \begin{tabular}{|c | c | c | c|}
            \hline
            \textbf{Dataset}  & \textbf{CUB~\cite{26_wah2011caltech}}  & \textbf{SUN~\cite{sun}} & \textbf{Aircraft-17~\cite{aircraft17}} \\
             \hline
             \textbf{Image Amount} & 11788 & 14340 & 1945 \\
             \hline
             \textbf{Categories} & 200 & 717 & 17 \\
             \hline
             \textbf{Attribute Amount} & 312 & 102 & 24 \\
             \hline
            \multirow{4}{*}{\textbf{Class Attributes}} &Bill Length & Running & Wing Shape \\
            &Tail Pattern & Dry & Aspect Ratio\\
            &Belly Color & Snow & Tail Type\\
            & Wing Shape & Leaves & Engine Number\\
            \hline
        \end{tabular}
    \end{center}
    \caption{The statistics of three datasets. We list four class attributes from CUB, SUN, and Aircraft-17.}
    \label{tab:datasets}
\end{table}

\subsection{Datasets.}
To verify the efficacy of EAT framework, we perform experiments on two fine-grained recognition datasets CUB~\cite{26_wah2011caltech}, Aircraft-17~\cite{aircraft17} and one large-scale scene recognition dataset SUN~\cite{sun}. Table ~\ref{tab:datasets} shows the statistics and some examples of the class attribute for three datasets.

CUB~\cite{26_wah2011caltech} is a dataset for bird classification, which contains $200$ categories with $11,788$ images. The dataset provides $28$ attribute groups and $312$ binary labels. All the attributes can be visually recognized in the image. The class attributes label is a matrix with size $200 \times 312$, 
    
Aircraft-17 is a aircraft classification dataset with $1,945$ images from $17$ categories~\cite{aircraft17}. The images are collected from Google Earth, with different resolutions ranging from $15 cm$ to $15 m$. The attributes for CUB dataset are publicly available, while the class attributes for Aircraft-17 are designed by ourself. To collect the attributes that distinguish each sub-category, we referred to the book ~\cite{endres2005jane} to define the attributes vocabulary. Firstly, we collect the vocabulary containing all the characteristics for 17 aircraft sub-categories. We select those attributes that can be viewed on the image, such as the wing and tail of the aircraft, instead of the semantic attributes, e.g., manufacturer and service date.

SUN~\cite{sun} is dataset for large-scale scene recognition, which consists of images covering a large variety of environmental scenes, places. The dataset contains $14,340$ images from $717$ scene categories. The author provided binary labels for $102$ class-attributes. However, some of the attributes are invisible on the image and should be inferred by other characters.
    
In our work, only class attributes are considered, so that all the images in one sub-category hold the same label for every attribute. This allows us to collect attribute labels for Aircraft-17 within one day.
Some respective attributes of CUB, SUN and Aircraft-17 are shown in Table~\ref{tab:datasets}. Despite the simplicity, the class attributes turn out to be very effective in the integrated classification process.

\subsection{Environmental Settings.}
The basic method is divided into two parts, ${{f}_{img}}(\cdot)$ and ${{g}_{c}}(\cdot)$. And ${{g}_{a}}$ is constructed following the structure of ${{g}_{c}}$. Both ${{g}_{c}}$ and ${{g}_{a}}$ are 4 layers CNN, and ${{g}_{i}}$ is 3 layers CNN. When calculating $c$ and $l$, $\lambda$ and $\eta $ take values between 0.5 and 1.5 for different datasets and basic methods. Our model is implemented using Pytorch~\cite{paszke2017automatic} and we plan to release our code to facilitate the reproduction of our results.

\subsection{Attention Map Generation}
\label{sec:Grad-CAM}
To reflect the model attention, we choose to use the visual explanation method Grad-CAM\cite{20_grad_cam}, which generates the Gradient-weighted Class Activation
Mapping. Grad-CAM has shown state-of-the-art attention localization ability in many evaluation methods~\cite{20_grad_cam,yang2019bim}, and we can easily apply Grad-CAM on existing CNN networks such as AlexNet~\cite{Alexnet} and ResNet~\cite{he2016deep}.

In order to generate the attention map, Grad-CAM back-propagate the gradients of the selected target score, and flow the gradients back to the last convolutional layer: 

\begin{equation}
    \alpha _{k}^{c}=\frac{1}{Z}\sum_{i}\sum_{j}{\frac{\partial y^c}{\partial A_{ij}^{k}}}
\end{equation}
where the weight $\alpha_c^k$ is calculated by the gradient of $y^c$ over the activation maps $A^k$, which reflects the importance of the last
convolutional layer’s activation map for that class. 

The attention map $AT$ is the weighted sum of all the activation maps $A^k$ in the last convolution layer
\begin{equation}
    AT = \mathit{ReLU}(\sum_{k}\alpha _{c}^{k}A^{k})\,.
\end{equation}

\subsection{Foreground Attention Rate}
\label{sec:FAR}
We propose a complementary metric, the Foreground Attention Rate, to quantitatively evaluate the effectiveness of EAT framework. As we know that the EAR framework will concentrate the network attention on the foreground objects. This effectiveness can be viewed by comparing two attention maps. However, given an attention map, we want to calculate the attention value focused on the forehead object quantitatively.
\begin{figure}
    \centering
    \includegraphics[width=0.4\textwidth]{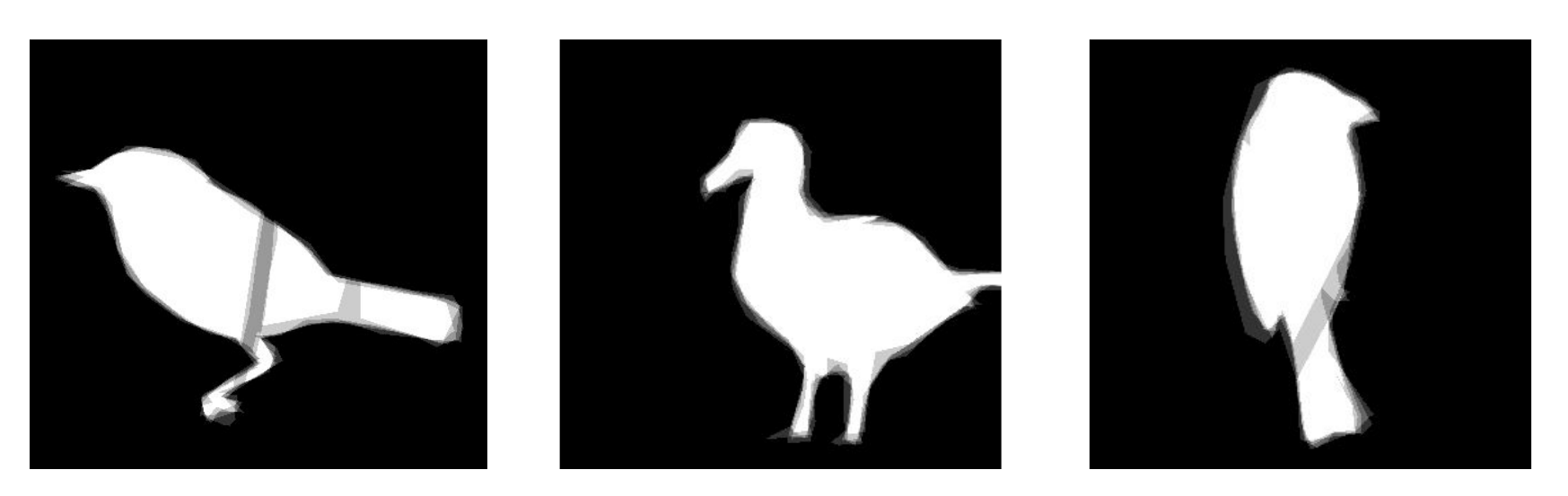}
    \caption{
            The masks for forehead object in CUB dataset. The pixels covering the foreground object have value one and others zero.
    } 
    \label{fig:foreground_object}    
\end{figure}

Given the binary segmentation mask $M$ for the forehead object (shown in Figure~\ref{fig:foreground_object}), where the pixels covering the object are one and others zero. One evaluation option is to directly accumulate the attention values that fall in the object mask $\sum{AT \odot M}$. However, a meaningless attention map that highlights all the pixels on the image, including object and background, would seem to perform well. Besides, two different models may generate various attention scale, i.e., a pixel with an attention value 0.5 might be essential in one model but not important in another model. Even if the model generates an attention map with a similar range of values, the mean and density of the values may be different.  Here we define Foreground Attention Rate (FAR) as an accurate indicator to measure the concentration of attention maps. The FAR rate is the ratio of averaged attention map value for a foreground object and the background:

\begin{equation}
FAR = \frac{PI(AT, M)}{PI(AT, (1-M))}
\end{equation}
where $PI(AT, *)$ is the average pixel importance of foreground object ($M$) and background ($1-M$), and $AT$ is the attention map we are evaluating. 

$PI(AT, M)$ is calculated as:

\begin{equation}
    PI(AT, M) = \frac{\left |AT\odot M \right |}{\left | M \right |},
\end{equation}
 where $\odot$ means element-wise multiplication.
 
The FAR rate provides an accurate and neutral evaluation for attention maps from different models. It will not be influenced by the density and scale of attention map. A higher FAR rate denotes that the network is paying more attention over object pixels. The experiment results for FAR rate is shown in Section~\ref{sec:FAR_result}

\section{Foreground Object Concentration}
\label{sec:foreground}
We perform image recognition tasks on three datasets, to quantitatively and qualitatively evaluate our EAT framework. We prove that attribute-based multi-task learning helps the model to concentrate on foreground objects and deal with the data bias problem. With the proposed Foreground Attention Rate, we can quantitatively prove our conclusion.

\subsection{Quantitative Analysis Among EAT Framework}

In this section, our EAT framework are trained with several basic models, AlexNet
~\cite{Alexnet}, ResNet18, ResNet50~\cite{he2016deep}, and PnasNet~\cite{liu2018progressive}.  Note that for fine-grained classification task, researchers propose various methods that achieve state-of-the-art accuracy~\cite{27_DFL,8099959, n5_li2018read, aircraft17}. However, our main task is not to improve the recognition accuracy. Thus we perform experiments on basic classification models.  To prove the EAT framework can be generalized on those well-designed methods, we combine EAT with one of the fine-grained classification network DFL~\cite{27_DFL,8099959} and gets the same trends as other basic models.

We investigate the change of classification accuracy after the model attention are concentrated on the object. Table~\ref{tab:cls-acc} shows the corresponding classification accuracy. The ``Original Category Prediction'' denotes the accuracy for the basic model without introducing class attributes. The ``Integrated Classification'' is the final accuracy of $c_i$ after introducing integrated classification.

 In CUB dataset, the performance of AlexNet, ResNet18, ResNet50 and PnasNet5 are increased by $2.72\%$, $2.23\%$, $0.69\%$ and $0.5\%$ respectively. In particular, EAT can also be generalized to state-of-the-art models. For instance, the performance of well-designed fine-grained classification method DFL~\cite{27_DFL} is improved by $0.35\%$. For Aircraft-17 dataset, the experimental results are similar. EAT improves the performance of AlexNet, ResNet18, RenNet50 and PnasNet5 by $1.53\%$, $1.08\%$, $0.94\%$ and $1.07\%$ in Aircraft-17. The $93.23\%$ accuracy achieved by PnasNet5+ABRM is higher than the state-of-the-art models FCFF ($91.38\%$) and SCFF ($93.15\%$)~\cite{aircraft17}. For scene recognition dataset SUN, EAT improves the performance of AlexNet, ResNet18, RenNet50 and PnasNet5 by $0.98\%$, $1.25\%$, $2.51\%$ and $2.33\%$. 

The results show a common trend that the classification accuracy is improved by two modules. It indicates that joint training the attribute prediction module and the original image category prediction will help to extract the subtle differences in the class attributes, and the classification accuracy is improved.

\begin{table}  
	\centering  
	\caption{The classification accuracy of various models. Here original category prediction denotes the performance of the base model without introducing class attributes. The integrated classification denotes the results of EAT framework.}
	\begin{tabular}{cc|c c}
						\toprule
						\multicolumn{1}{c|} {\multirow{2}{*}{\textbf{Dataset}}} & \multirow{2}{*}{\textbf{Base model}} &\textbf{Original Category} & \textbf{Integrated} \\
						\multicolumn{1}{c|}{}&& \textbf{Prediction} & \textbf{Classification} \\
						\midrule
						\multicolumn{1}{c|}{\multirow{2}{*}{\textbf{}}}&Alexnet~\cite{Alexnet} & 68.33\%& \textbf{71.05\%} \\
						\multicolumn{1}{c|}{\multirow{2}{*}{\textbf{CUB~\cite{26_wah2011caltech}}}}&Resnet18~\cite{he2016deep} & 75.35\%& \textbf{77.58\%}\\
						
					    \multicolumn{1}{c|}{}&ResNet50~\cite{he2016deep} & 83.27\%& \textbf{83.96\%}\\
						\multicolumn{1}{c|}{}&PnasNet5~\cite{liu2018progressive} & 84.60\%& \textbf{85.10\%}\\
						
						\multicolumn{1}{c|}{}&DFL~\cite{27_DFL} &85.82\% & \textbf{86.17\%} \\
						\midrule
						\multicolumn{1}{c|}{\textbf{}}&AlexNet~\cite{Alexnet} & 85.43\%  & \textbf{86.96\% }\\
						\multicolumn{1}{c|}{\textbf{Aircraft-17~\cite{aircraft17}}}&ResNet18~\cite{he2016deep}  & 89.87\% & \textbf{90.18\%}\\
						
						\multicolumn{1}{c|}{}&ResNet50~\cite{he2016deep} & 91.64\%& \textbf{92.58\%}\\
						\multicolumn{1}{c|}{}&PnasNet5~\cite{liu2018progressive} & 92.16\%  & \textbf{93.23\%} \\
						
						\multicolumn{1}{c|}{}&DFL~\cite{27_DFL} &92.05\% & \textbf{92.90}\% \\
						
						\midrule
						
						
						\multicolumn{1}{c|}{\multirow{2}{*}{\textbf{}}}&Alexnet~\cite{Alexnet} & 37.97\%& \textbf{38.95\%} \\
						\multicolumn{1}{c|}{\multirow{2}{*}{\textbf{SUN~\cite{sun}}}}&Resnet18~\cite{he2016deep} & 40.28\%& \textbf{41.53\%}\\
						
					    \multicolumn{1}{c|}{}&ResNet50~\cite{he2016deep} & 42.47\%& \textbf{44.98\%}\\
						\multicolumn{1}{c|}{}&PnasNet5~\cite{liu2018progressive} & 42.71\%& \textbf{45.04\%}\\
						
						\bottomrule
					\end{tabular}
	\label{tab:cls-acc}  
\end{table}  
The average attribute classification accuracy is shown in Table~\ref{tab:attributes_acc}. For CUB and Aircraft-17, the accuracy for attribute classification is high. For SUN, the accuracy is relatively lower. It is because all the attributes in CUB and Aircraft-17 can be viewed on the image, such as the shape, color or number. However, some of the attributes in SUN cannot be directly inferred from the image, e.g., ``business'' and ``smoothing''. In order to infer the value of those attributes, the network needs to predict the object category first. No visual evidence can be found on the image. Thus the accuracy for these attributes is relatively low.

\begin{table*}  
	\centering  
	\caption{The average attribute classification accuracy for three datasets.}
		\begin{tabular}{c|c c c c c}
			\toprule
			Dataset/Model & AlexNet~\cite{Alexnet} & ResNet18~\cite{he2016deep}& ResNet50~\cite{he2016deep}& PnasNet5~\cite{liu2018progressive}& DFL~\cite{27_DFL}\\
			\midrule
			CUB & $89.21\%$ & $90.35\%$ & $91.15\%$ & $92.69\%$ & $92.58\%$\\
			\midrule
			SUN & $83.19\%$ & $83.87\%$ & $85.82\%$ & $86.10\%$ & $-$\\
			\midrule
			Aircraft-17 & $90.33\%$ & $92.32\%$ & $92.97\%$ & $93.46\%$ & $95.51\%$\\
			\bottomrule
		\end{tabular}
	\label{tab:attributes_acc}  
\end{table*}  


\subsection{Qualitative Analysis}

In order to investigate and compare the attention of the original basic model and the EAT framework, we generate their attention maps for category prediction. As described in Section~\ref{sec:Grad-CAM}, we use Grad-CAM to generate attention map. The basic model in the EAR framework is ResNet50~\cite{he2016deep}.

\begin{figure}
    \centering
    \includegraphics[width=0.5\textwidth]{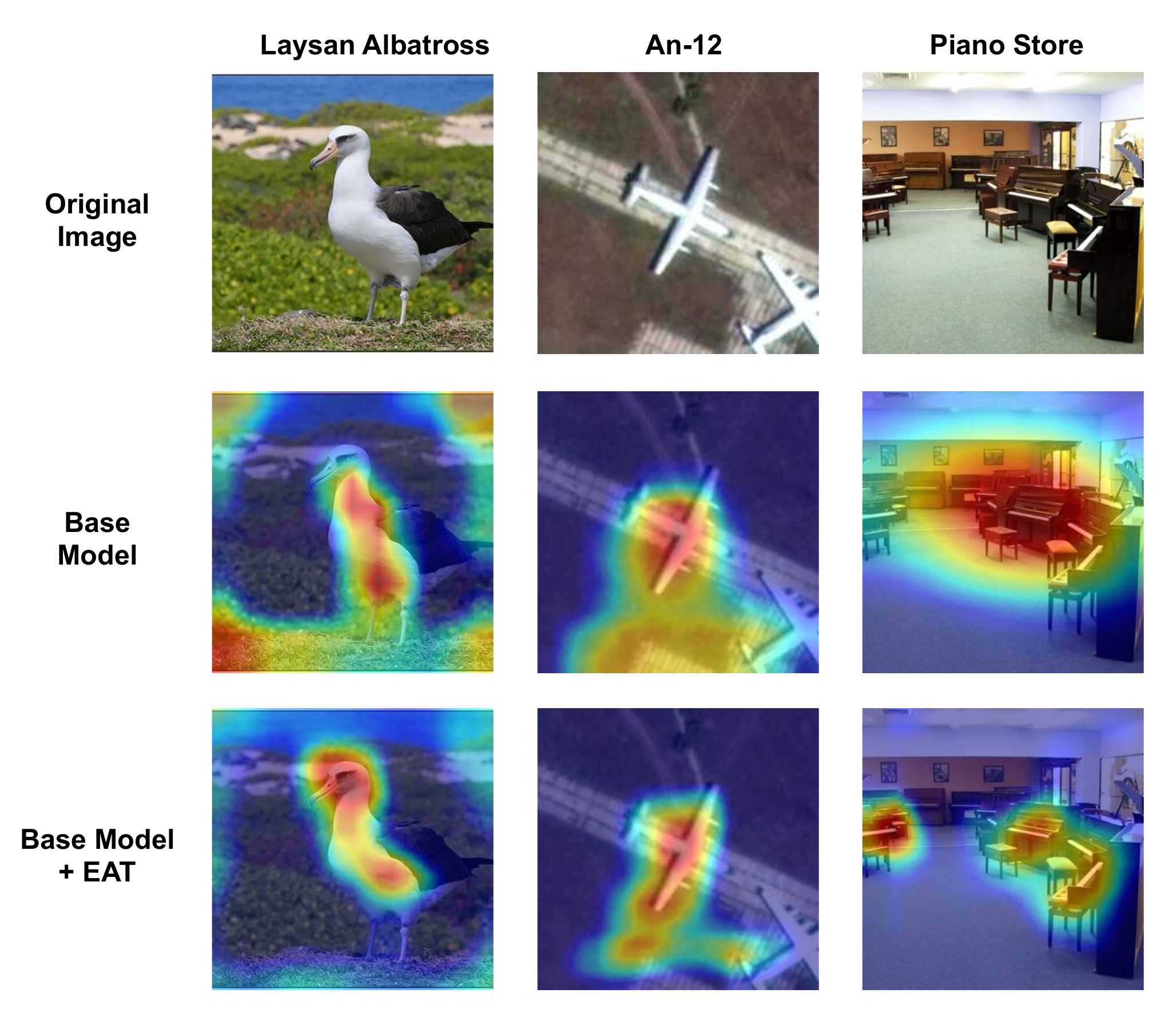}
    \caption{
            The attention maps for the category classification. The first row lists original images. The second row is the classifier's attention maps of the original model without joint training with attribute prediction. The third row is the attention map of the model in the EAT framework.
    } 
    \label{fig:forehead}    
\end{figure}

Figure~\ref{fig:forehead} shows some of the attention maps. The first row is the original image, and the second row is the attention of the base model without introducing class attributes $AT_o$. The third row is the attention of the final category prediction in EAT framework $AT_e$. In the second row, although most of the attention lies on the foreground object, there is still a notable part of the attention that falls on the background. For instance, when classifying the bird ``Laysan Albatross'', the model is disturbed by the image edge and leaves in the background. This phenomenon is caused by the dataset bias: nearly all the birds in the dataset are accompanied by their habitats, such as forest and ocean. Similarly, when classifying the aircraft, the background around the aircraft, such as the runway and the apron, affects the model's attention. 

When training the network for a single task, the image category labels cannot provide more discriminative information to help the model from overfitting. Thus the network tries to find all the evidence that can map the image to the category label. The background is regarded as evidence for classification. In contrast, since the background information is the same for all the attributes, thus it cannot provide any evidence when classifying attributes. In our EAT framework, image category prediction and attribute classification share the basic model. Thus the parameters are trained to find the optimal solution for two tasks. When the network is jointly trained for attribute classification, it is less prone to overfit on the original task.

The result is proved in the third row of Figure~\ref{fig:forehead}. After applying EAT framework to the basic model, the focus on the background and image edges are significantly reduced.
Besides, more attention is focused on the subtle properties of the foreground object, e.g.\ the attention accurately covers the pianos in the store. We also observe that some attention moved from the less discriminative parts to the prominent parts. For instance, the attention maps are focused on the hooked bill and white belly of ``Laysan Albatross'', and the outline of the airplane ``An-12'' is masked out. Since the shared layers are updated together during the training process, the attribute prediction helps the model to tell which characteristics of the target are more distinguishable.

\subsection{FAR Rate Analysis }
\label{sec:FAR_result}

To calculate how much does the network concentrate attention on the foreground objects, we define Foreground Attention Rate as an indicator. We perform experiments to quantitatively compare the attention map generated by the original basic model and by the EAT framework.

Table~\ref{tab:FARate} shows the FAR rate on CUB dataset. The FAR rate for five basic models is between $3$ to $4$. That is to say, the average attention value on object pixels is three to four times larger than that of background pixels. In the EAT framework, the FAR rate is improved by nearly $60\%$. It indicates that after applying the EAT framework, the attention is moved from background to the forehead object. To the best of our knowledge, this is the first quantitative evaluation for the concentration of attention maps.

\begin{table*}  
    \centering  
    \caption{The Foreground Attention Rate (FAR) rate. ``Base Model'' is the original model without joint training with attribute prediction. ``Base Model + EAT'' is the model in EAT framework. A higher rate means more attention is focused on the foreground object.}
        \begin{tabular}{c|c c c c c}
            Models & AlexNet~\cite{Alexnet} & ResNet18~\cite{he2016deep}& ResNet50~\cite{he2016deep}& PnasNet5~\cite{liu2018progressive}& DFL~\cite{27_DFL}\\
            \toprule
            Base Model & 3.36 & 3.44 & 3.70 & 3.95 & 3.89\\
            \midrule
            Base Model + EAT & 5.27 & 5.39 & 6.04 & 6.35 & 6.21\\
            \bottomrule
        \end{tabular}
    \label{tab:FARate}  
\end{table*}  


\section{Attribute-based Explanation}

In this section, we show the attribute-based explanations generated by our framework. The attributes in the explanation can also be grounded on the images with attention map, giving a multi-model interpretation. We also analyze the explanations for correct and wrong predictions, to reveal how can they help users in understanding and improving the model and data.

\subsection{Multi-model Explanations}

\begin{figure*}
    \centering
    \includegraphics[width=1\textwidth]{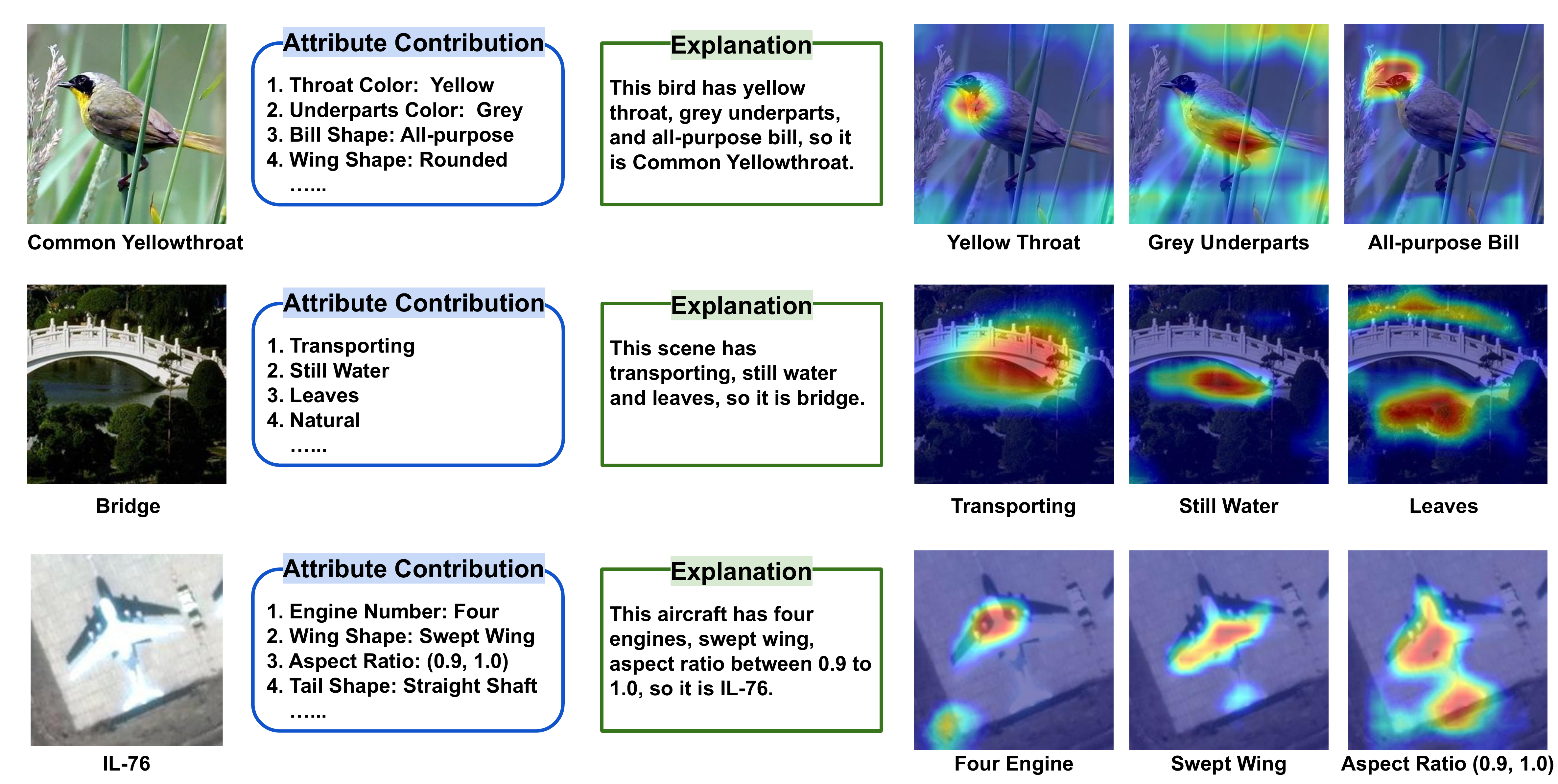}
    \caption{
            The results of the Embedding Attention Reasoning. From left to right, we list the original image, the rank of attributes with the high contribution in EAR module, the explanation for the classification, and the attention map for attributes in the explanation. We choose the top-3 attributes with the highest contribution to form the attribute-based explanation. We can also visually ground the explanation on the image.  The attention maps are generated by backpropagating from attributes predicted results $c_a$.
    } 
    \label{fig:explanation_right}    
\end{figure*}

In this section, we discover the critical attributes that can distinguish one image from other categories. Then the top-3 attributes with the highest contribution value are summarized into a textual explanation, to indicate what helps the model to make a decision. In Figure~\ref{fig:explanation_right}, we show the important attributes and explanations. For instance, the attribute ``Throat Color: Yellow'' has the greatest influence for the decision of ``Common Yellowthroat'', and ``Transporting'' has the strongest influence for the decision of ``Bridge''. 

Despite the textual explanations, we can also visually ground the attributes in the explanation on the image with attention maps. We backpropagate the results of attributes prediction to the input image and generate the attention maps. The attention maps highlight the image region that the network is focusing on when classifying the attributes. In this way, we provide a visual and textual combined interpretation for every image.

Figure~\ref{fig:explanation_right} shows the visual localization results for attributes. Compared to the attention map for the species prediction (in Figure~\ref{fig:foreground_object}), the attribute attention maps provide more discriminative information. As we can see, the Attribute Prediction Module can focus on the corresponding parts of the attributes. When classifying the Bridge, the attention for ``Transporting'', ``Leaves'' and ``Still Water'' are on the corresponding parts. When predicting the ``Bill Shape'' and ``Underparts Color'' of the bird, the network attention covers the head and belly respectively. The visual grounding for attributes can be generated together with the textual explanations. They form a multi-modal interpretation that can provide more accurate information.

While some fail cases focus on wrong parts when predicting an attribute. The classifiers for ``IL-76'' can accurately locate the position of ``Swept Wing'' and ``Engine'', but the attention map for ``Aspect Ratio'' are distracted by the tail. This is because the ``Aspect Ratio'' is an abstract attribute that the network cannot understand. When the visual grounding is not accurate, it can also reveal the network weakness.

\subsection{Interpretation for Correct Predictions}

For those correct predictions, the explanations state the most discriminative characters that influence the network decision and give users a better understanding of the classification network. In Figure~\ref{fig:explanation_right}, the reason for classifying the bird ``Common Yellowthroat'' lies in ``the yellow throat, grey underparts and the all-purpose bill''. To classify the bridge, the ``transporting property, the leaves around the bridge and the still water'' helps the network to make the right decision. The reason for classifying the aircraft ``KC-10'' lies in ``the four engines, the swept wing, and the aspect ratio''. 

Besides, inexperienced users would not know how to discriminate the subspecies in the fine-grained recognition task. For instance, an ordinary user can tell a bird from other animals, but cannot tell which subspecies does the bird belong to, the ``Fox Sparrow'' or ``Song Sparrow''.  Thus the correct explanations can work as a machine teaching tool. By stating the essential parts that influence network decision, the explanations can help users to tell the difference between adjacent species. As is shown in Figure~\ref{fig:explanation_wrong}, the explanation for the second image states the most discriminative difference, the white back, between ``Sooty Albatross'' and ``Black Footed Albatross''. The explanation for the five image highlights the ``grey foot'' that can distinguish ``Baltimore Oriole'' from ``American Goldfinch''. These interpretations can be good examples to teach the user.

\subsection{Interpretation for Wrong Predictions}
\begin{figure*}
        \centering
        \includegraphics[width=1\linewidth]{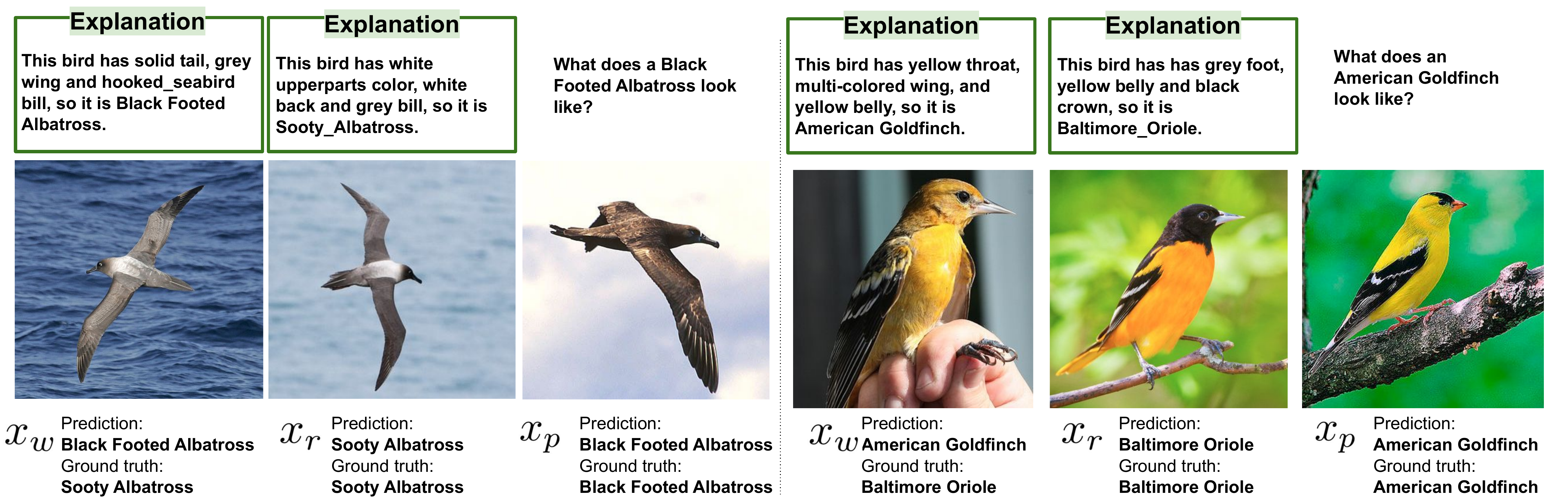}
        \caption{ The explanation for the wrong predicted images $x_w$, a correctly predicted image $x_r$ from $x_w$'s ground truth category, and an example $x_p$ from $x_w$'s predict category. We listed the explanation for wrongly predicted image $x_w$, and correctly predicted image $x_r$.
        }
        \label{fig:explanation_wrong}    
\end{figure*}

Except for checking the explanations for right predictions, investigation of wrong cases would give more information towards the problem of the network.

In Figure~\ref{fig:explanation_wrong}, we select some wrong cases from the CUB dataset. The experiment is performed on EAT framework with ResNet50, which achieves a classification accuracy of $84.66\%$. Here we show the explanation for the wrong predicted images $x_w$, a correctly predicted image $x_r$ from the same category as $x_w$, and an example $x_p$ from the class that $x_w$ is classified to. As is shown in the figure, $x_w$ and $x_p$ looks similar. $x_w$ share some commonality with $x_p$ that misleads the network to make wrong predictions. The reason for the wrong prediction can be various, and only show the images cannot point to the evidence. While the explanations would reveal the reason for those errors. 

For instance, the model might focus on the commonality between $x_w$ and $x_p$, while ignoring their differences. In Figure~\ref{fig:explanation_wrong}, the first bird ``Sooty Albatross'' is wrongly classified to ``Black Footed Albatross''. The explanation indicates that the network lays much importance on the tail, wing and bill, which are quite similar to that of ``Black Footed Albatross''.

Other reasons might be the defect of the testing data. Sometimes the testing data is not typical. Thus it would be hard to recognize them. As is shown in Figure~\ref{fig:explanation_wrong}, the $x_w$ of Baltimore-Oriole shares the same commonalities with $x_r$ and $x_p$, such as the yellow throat, multi-colored wing and the yellow belly. However, the most discriminative attribute, the black head, are not visible in $x_w$. Thus it is hard to be recognized, and the model should not be blamed on making a mistake. In overall, analyzing the wrong cases can provide constructive information for improving the model as well as the data.


\section{Conclusion}
In this paper, we propose an explainable attribute-based multi-task framework (EAT) to performs image classification and attribute recognition together. The joint training of these two tasks helps the model to focus on the foreground objects, and deal with the data bias problem. We propose an embedding attention reasoning module to investigate which attributes contribute more to the classification result. The investigation on attributes allows us to generate attribute-based explanation. The important attributes can also be visually grounded on the image via model attention, providing multi-modal explanations for the decision process. We perform experiments on five basic models and three datasets. Results show that our framework achieves significant improvement over these models. 

As shown in the experiment, our framework can be generalized to the state-of-the-art methods and further improve the performance. The attribute-based explanation can discriminate the adjacent species. Thus we can use explanations to teach unexperienced users about the difference between sub-species.

Our experiments also reveal that the quality of attributes will influence the performance of EAT framework. In future work, we aim at creating more conductive and accurate attributes. The attributes should be visible on the image and be discriminative between species.




\bibliography{mybib}

\end{document}